\documentclass[runningheads]{llncs}
\usepackage{graphicx}

\usepackage{tikz}
\usepackage{comment}
\usepackage{amsmath,amssymb,bm} 
\usepackage{color}
\usepackage{multirow}
\usepackage{wrapfig}
\usepackage{bbding}
\usepackage{booktabs}

\usepackage[accsupp]{axessibility}  

\usepackage[hidelinks,colorlinks=true]{hyperref}

\begin{document}
\pagestyle{headings}
\mainmatter

\title{Rethinking IoU-based Optimization for Single-stage 3D Object Detection} 

%
\author{Hualian Sheng\inst{1,2,3}\and
Sijia Cai\inst{2} \and
Na Zhao\inst{3}\thanks{Corresponding author} \and Bing Deng\inst{2} \and Jianqiang Huang\inst{2} \and Xian-Sheng Hua\inst{2} \and Min-Jian Zhao\inst{1} \and Gim Hee Lee\index{Lee, Gim Hee}\index{Lee, Gim Hee}\inst{3}}
\authorrunning{H. Sheng et al.}
%
\institute{College of Information Science and Electronic Engineering, Zhejiang University\\
\and Alibaba Cloud Computing Ltd.\\  \and 
Department of Computer Science, National University of Singapore\\  
\email{\{hlsheng,mjzhao\}@zju.edu.cn}, \email{\{nazhao, gimhee.lee\}@comp.nus.edu.sg},\\
\email{\{stephen.csj, dengbing.db, jianqiang.hjq, xiansheng.hxs\}@alibaba-inc.com}
}
\maketitle

\begin{abstract}
	Since Intersection-over-Union (IoU) based optimization maintains the consistency of the final IoU prediction metric and losses, it has been widely used in both regression and classification branches of single-stage 2D object detectors. 
	Recently, several 3D object detection methods adopt IoU-based optimization and directly replace the 2D IoU with 3D IoU. However, such a direct computation in 3D 
	is very costly due to the complex implementation and inefficient backward operations. 
	Moreover, 3D IoU-based optimization is sub-optimal as it is sensitive to rotation and thus 
	can cause training instability and detection performance deterioration. 
	In this paper, we propose a novel Rotation-Decoupled IoU (RDIoU) method that can mitigate the rotation-sensitivity issue, and produce more efficient optimization objectives compared with 
	3D IoU during the training stage. Specifically, our RDIoU simplifies the complex interactions of regression parameters by decoupling the rotation variable as an independent term, yet preserving the geometry of 3D IoU. By incorporating RDIoU into both the regression and classification branches, 
	the network is encouraged to learn more precise bounding boxes and concurrently overcome the misalignment issue between classification and regression. Extensive experiments on the 
	benchmark KITTI and Waymo Open Dataset validate that our RDIoU method can bring substantial improvement for the single-stage 3D object detection. The code is available at \href{https://github.com/hlsheng1/RDIoU}{https://github.com/hlsheng1/RDIoU}.
\end{abstract}

\begin{figure}[t]
	\centering
	\includegraphics[width=0.88\textwidth]{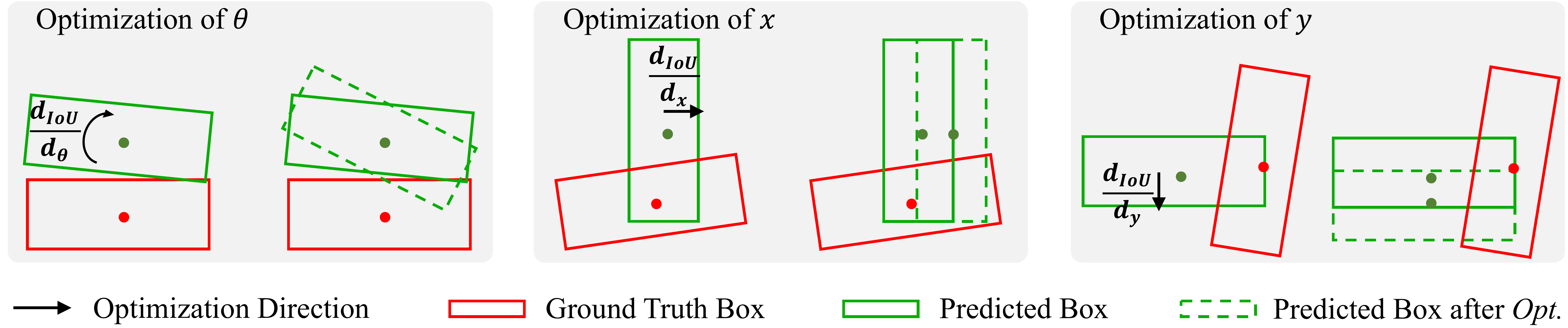} 
	\caption{Illustration of direct 3D IoU optimization 
	between the ground truth box (red) and predicted box (green) in Bird’s Eye View (BEV), compared to 
	the post-optimized bounding box (dotted green).
	We see that optimizing the 3D IoU over the bounding box parameters does not bring the predicted boxes closer to the ground truths. 
	}
	\label{Real_IoU}
\end{figure}

\section{Introduction}
Point cloud-based 3D object detection is an important task in robotics and autonomous driving.
In contrast to the image-based object detection task, 3D object detection needs to predict 3D bounding boxes with 
higher degrees-of-freedom and allocate confidence map in a larger search space. This 
causes difficulties in predicting accurate 3D bounding boxes with reliable confidence.

State-of-the-art 3D detection approaches can be generally divided into single-stage
and two-stage methods. Two-stage methods~\cite{chen2017multi,qi2018frustum,shi2020pv,shi2019pointrcnn,yin2021center} usually perform a second feature extraction or multi-scale feature aggregation based on the proposals generated by the region proposal network (RPN). As a result, the 
second-stage network can focus on the partial positive areas to avoid the sparsity of the whole point cloud. Furthermore, the second-stage network can utilize Intersection-over-Union (IoU) guided supervision to generate more reliable confidence predictions.
In contrast to the coarse-to-fine pipeline in the two-stage methods, single-stage methods~\cite{he2020structure,lang2019pointpillars,yan2018second,yin2021center,zheng2020cia,zheng2021se} adopt an end-to-end pipeline that designs one dense object detector to directly predict pixel-level object categories and bounding boxes. The end-to-end one-stage pipeline is much more elegant and effective than the two-stage pipeline. Nevertheless, the performance of single-stage methods is usually inferior to the two-stage methods. The performance gap is mainly caused by inaccurate regression precision and confidence assignment of the single-stage methods since they independently predict the 3D bounding box center, size and rotation with the Smooth-$\ell$1 loss~\cite{girshick2015fast}, which results in inconsistency between the loss and the final IoU-based evaluation metric.

To maintain the consistency between the loss and IoU-based evaluation metric, IoU-based optimization~\cite{li2021generalized,li2020generalized,rezatofighi2019generalized,zheng2020distance}
has been widely studied and shown impressive performance improvements in image-based single-stage object detection. Inspired by this, several single-stage 3D object detection approaches~\cite{zheng2021se,zhou2019iou} adopt the IoU-based optimization by replacing the 2D IoU with the 
3D IoU. Despite their impressive performance gain in the axis-aligned cases, the computation of intersection area between two rotated 3D bounding boxes is much more complex than their 2D counterparts without rotation. A number of efforts~\cite{chen2020piou,yang2021rethinking,yang2019scrdet,zheng2020rotation} have been made to simplify the computation of rotated IoU intersection by approximations. 
For example, PIoU~\cite{chen2020piou} 
only counts the pixels in the intersection area. 
However, none of these methods can be directly utilized to solve the \textbf{negative coupling effect of rotation on the 3D IoU}. 
As illustrated in Figure~\ref{Real_IoU}, 
optimizing the 3D IoU over the bounding box parameters can lead to further misalignment between the predicted and ground truth bounding boxes. This problem is caused by the coupling between the rotation and the center and size parameters of the 3D bounding box. In contrast, the 
issue of negative coupling effect of rotation does not exist in the 2D IoU counterpart under common settings as the optimization over the center and size parameters directly minimizes the 2D IoU without rotation.
To this end, we propose a new high-performance and efficient 3D bounding box optimization
objective that 
satisfies the following three conditions: it must 1) be differentiable; 2) satisfy the consistency between the evaluation metric and optimization objective; 3) circumvent the negative coupling effect of rotation on the 3D IoU. 

In this paper, we propose a Rotation-Decoupled IoU (RDIoU) method to model the interaction of two arbitrary-oriented 3D bounding boxes while stabilize the training process. Our main idea is to disentangle dependencies of the bounding box parameters by decoupling and handling the rotation individually at a loss level. Specifically, we utilize a 4-dimensional 
representation of the box to calculate the IoU-like criterion and remove complex rotation-edge variations. A fixed hyperparameter is introduced to further control the weights of rotation change in the computation of our RDIoU. Our proposed RDIoU formulation addresses the non-differentiable and instability issues caused by the rotation change while preserving the geometry 
of 3D IoU.
By incorporating RDIoU into the regression and classification branches, we propose a RDIoU-guided DIoU loss~\cite{zheng2020distance} and a RDIoU-guided quality focal loss~\cite{li2021generalized,li2020generalized}. 
These two enhanced losses guide the optimization towards high-performance box localization and alleviate the misalignment between classification and regression, and thus significantly improve the single-stage 3D detection performance over other existing losses.

Based on our proposed RDIoU, we build a simple and elegant single-stage 3D detector. In the training phase, our RDIoU-based optimization guides the 3D CNN backbone to achieve better feature alignment and get more accurate box parameters without encountering the rotation-sensitive issues. Furthermore, our RDIoU-based single-stage 3D detector also enables us to effectively train the entire
network in an end-to-end manner without the need of hyperparameter-sensitive, stage-wise training or time-consuming backward operations. Our main contributions in this paper are: 1) We propose the RDIoU-based optimization for 3D object detectors, which brings a more robust training process as compared to the 3D IoU-based optimization strategy. 2) We incorporate RDIoU into regression supervision and present the RDIoU-guided DIoU loss, which exhibits significantly better performance as compared to the 3D IoU-guided DIoU loss~\cite{zheng2021se}. 3) We incorporate RDIoU into classification supervision and present the RDIoU-guided quality focal loss, which is able to help generate more reasonable confidence maps. 4) We conduct extensive experiments on two benchmark datasets (\textit{i.e.,} KITTI and Waymo), and the promising results validate that our RDIoU can attain top performance with the commonly used backbone networks.

\section{Related Work}

\noindent{\textbf{Single-stage Detectors.}} Single-stage detectors predict the location and category from predefined anchor boxes or points over different spatial positions in a single-shot manner such as SECOND~\cite{yan2018second} and PointPillar~\cite{lang2019pointpillars}. SECOND~\cite{yan2018second} is the first to apply 3D sparse convolution~\cite{graham2015sparse,graham2017submanifold} to process the 3D voxel features, which are averaged by voxel-enclosed points. PointPillar~\cite{lang2019pointpillars} collapses the points in vertical pillars with a simplified PointNet~\cite{qi2017pointnet,qi2017pointnet++}, followed by a typical 2D CNN backbone. Point-GNN~\cite{shi2020pointgnn} proposes a graph neural network to learn better point features. 3DSSD~\cite{yang20203Dssd} develops F-FPS as a supplement of D-FPS to build an anchor-free 3D object detector. SA-SSD~\cite{he2020structure} applies an auxiliary segment network to assist the voxel feature learning. CIA-SSD~\cite{zheng2020cia} utilizes an IoU-aware confidence rectification to improve the classification. SE-SSD~\cite{zheng2021se} proposes a self-ensembling training schedule to improve the performance of a pretrained CIA-SSD model. These single-stage detectors are usually of high speed. 

\noindent{\textbf{Two-stage Detectors.}} Two-stage detectors can be seen as the extension of single-stage detectors. These methods first generate high-quality proposals with categories based on RPN, and then refine each proposal's location and output the prediction confidence. PointRCNN~\cite{shi2019pointrcnn} generates proposals based on segmented foreground objects and then refines the bounding box via a regression branch. Part $A^2$~\cite{shi2020points} extends PointRCNN with an intra-object part supervision. PV-RCNN~\cite{shi2020pv} first utilizes voxel-based backbone to generate high-quality proposals, and then devotes a point-based approach to aggregate multi-scale voxel features along with raw point features. Its variant Voxel-RCNN~\cite{deng2020voxel} proposes an accelerated point-based module to capture multi-scale voxel features without raw point features. LiDAR-RCNN~\cite{li2021lidar} introduces a plug-and-play point-based module to refine the RPN proposals based on simple PointNet-like approach. CenterPoint~\cite{yin2021center} suggests an anchor-free method to improve the RPN detect head, and then learns more accurate bounding box regression based on BEV features within given proposals. Transformer architecture~\cite{vaswani2017attention} recently appears in the 3D detection area. VoTr~\cite{mao2021voxel} improves the region proposal network by introducing Transformer into sparse convolution. CT3D~\cite{sheng2021improving} proposes a novel channel-wise Transformer for proposal refinement. These two-stage detectors usually exhibit high performance but suffer the 
problem of having too many hyperparameters and high latency, and thus 
limiting their usefulness in the industrial application. In comparison, it is 
imperative to improve the single-stage detector for the better trade-off between performance and latency.

\noindent{\textbf{IoU-based Optimization.}} IoU-based optimization has been effectively validated and implemented in 2D object detection without rotation, e.g. GIoU loss~\cite{rezatofighi2019generalized}, DIoU/CIoU loss~\cite{zheng2020distance}, quality focal loss~\cite{li2021generalized,li2020generalized}, etc. These methods aim to directly optimize the final evaluation metric and have achieved promising performance. 
Recently, several 3D rotated object detection works study a direct extension of the 2D IoU to 3D domain and yield the 3D IoU loss~\cite{zhou2019iou} and ODIoU loss~\cite{zheng2021se}. However, these 3D IoU-based losses usually need high resource cost such as huge GPU memory cost due to brute-force search or time-consuming back-propagation on the CPU.
Moreover, it is sub-optimal to set the 3D IoU as the 
optimization objective as shown in Figure~\ref{Real_IoU}. Instead, we propose a more effective optimization objective, \textit{i.e.} the RDIoU, to achieve both high-performance training and 
efficient back-propagation on the GPU.

\section{Method}
Our main contribution is the RDIoU, which 
is a better optimization 
objective than the 
existing 3D IoU in alleviating the negative coupling effect of the rotation during the training stage. In the following, we first discuss the details of our RDIoU and the comparison with 
3D IoU. Subsequently, we insert our RDIoU into the regression and classification branches of existing 3D object detectors. 

\begin{figure}[t]
	\centering
	\includegraphics[width=1\textwidth]{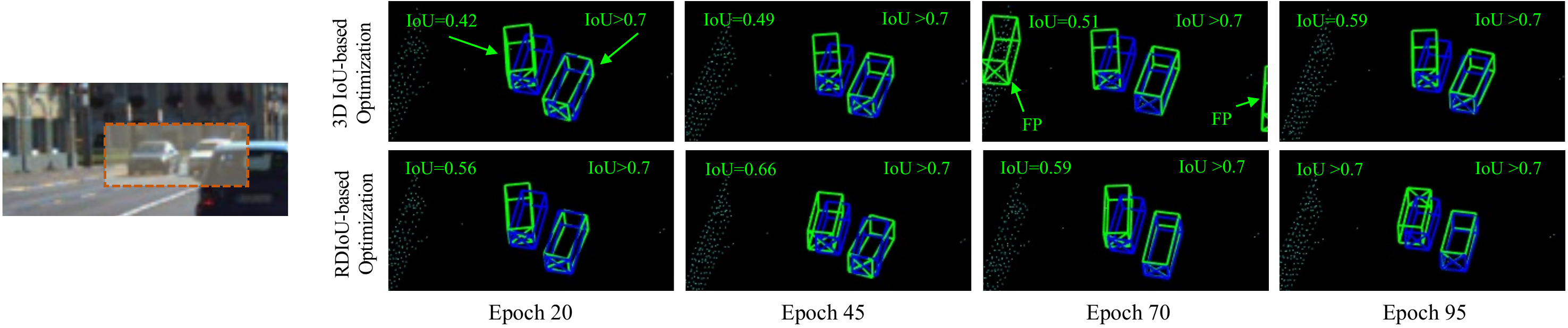} 
	\caption{Qualitative comparison between the 3D IoU-based and RDIoU-based optimization strategies on the KITTI \textit{train} set. The predicted and ground truth bounding boxes are shown in green and blue, respectively. The 3D IoU-based strategy is difficult to handle the hard sample, and the IoU is eventually stagnant under 0.7. In contrast, RDIoU-based strategy finally makes the prediction much closer to the ground truth.}
	\label{visrdiou}
\end{figure}

\subsection{Our RDIoU}\label{Sec_RDIoU}
The optimization strategies for object detection generally fall into two types: 
\begin{enumerate}
	\item{Non IoU-based: Each box parameter is individually optimized without considering their spatial connection. Typically, Smooth-$\ell 1$ loss~\cite{girshick2015fast} and focal loss~\cite{lin2017focal} are used for regression and classification, respectively.} 
	\item{IoU-based: The box parameters are jointly optimized via maintaining consistency with the IoU evaluation metric. Typically, DIoU/CIoU loss~\cite{zheng2020distance} and quality focal loss~\cite{li2021generalized} are used for regression and classification, respectively.}
\end{enumerate}

\begin{figure*}[t]
	\centering
	\includegraphics[width=1.0\textwidth]{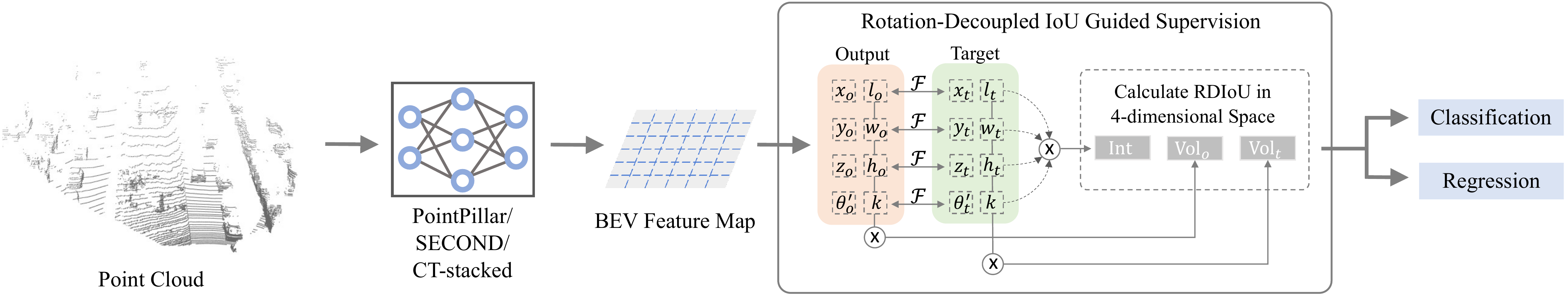}
	\caption{The illustration of RDIoU method. We advocate a Rotation-Decoupled IoU to model the intersection of two rotated 3D bounding boxes. This design is further set as the optimization target to guide the classification and regression learning, respectively.} 
	\label{model}
\end{figure*}

Compared to non IoU-based optimization, the IoU-based optimization strategy usually exhibits better performance due to the end-to-end metric learning. However, a naive  application of this strategy on 3D object detection causes potential instability issue in the optimization due to the negative coupling effect of the rotation and thus affects the final convergence of the network, as shown in the first row of Figure~\ref{visrdiou}. 
In view of this problem, we construct a more efficient optimization strategy that considers 
both the stability of the training process and 
the preservation of the geometry from the IoU on the bounding box. Figure~\ref{model} illustrates our proposed RDIoU method.

Let $(x_o,y_o,z_o,l_o,w_o,h_o,\theta_o)$ denotes the 3D bounding box parameters of the regression vector from the regression branch and $(x_{t},y_{t},z_{t},l_{t},w_{t},h_{t},\theta_{t})$ denotes its 
corresponding target regression vector. $(x, y, z)$ is the center coordinate, $(l,w,h)$ is the size, and $\theta$ is the rotation of the bounding box. Furthermore, we use $(x_a,y_a,z_a,l_a,w_a,h_a,\theta_a)$ to denote the selected anchor box from the 3D object detector and $(x_g,y_g,z_g,l_g,w_g,h_g,\theta_g)$ to denote the ground truth 3D bounding box. Following the previous works~\cite{yan2018second}, the regression target can be encoded as:
\begin{align}
	&x_{t} = \frac{x_g-x_a}{d}, y_{t} = \frac{y_g-y_a}{d}, z_{t} = \frac{z_g-z_a}{h_a}, \label{label_trans_1}\\
	&l_t = \frac{l_g}{l_a}, \quad w_t = \frac{w_g}{w_a}, \quad h_t = \frac{h_g}{h_a}, \quad  \theta_t = \theta_{g} - \theta_{a},  \label{label_trans_2}
\end{align}
where $d = \sqrt{(l_a)^2 + (w_a)^2}$ is the diagonal of the base of anchor. 

Since the negative coupling effect of rotation is the main source of inaccuracies in the 3D IoU-based optimization, we decouple the rotation into an independent dimension to avoid its complex coupling relationship with the other bounding box parameters. Specifically, the rotation is decoupled into the 4-$th$ dimension with a fixed edge $k$. 
As a result, the center of the optimized bounding box is transformed from $(x_t,y_t,z_t)$ to $(x_t,y_t,z_t,\theta_t)$, and thus we can easily calculate the IoU in this 4-dimensional space. Consequently, the RDIoU of two rotated 3D bounding boxes can be formulated with a regular IoU~\cite{rezatofighi2019generalized} as:
\begin{equation}
	\text{RDIoU} = \text{Int}/(\text{Vol}_{o}+\text{Vol}_{t}-\text{Int}),\ \  \text{where}
\end{equation}
\begin{equation*}
\begin{split}
	 &\text{Int}= \mathcal{F}(x_o,x_{t},l_o,l_{t})*\mathcal{F}(y_o,y_{t},w_o,w_{t})*\mathcal{F}(z_o,z_{t},h_o,h_{t})*\mathcal{F}(\theta_{o'},\theta_{t'},k,k),\\
	 &\text{Vol}_{o} = l_{o}*w_{o}*h_{o}*k,~~~\text{Vol}_{t} = l_{t}*w_{t}*h_{t}*k, \\ \text{and} \quad & \mathcal{F}(a_o,a_{t},b_o,b_{t}) = \min(a_o+\frac{b_o}{2},a_{t}+\frac{b_{t}}{2})-\max(a_o-\frac{b_o}{2},a_{t}-\frac{b_{t}}{2}). 
 \end{split}
\end{equation*}
Following the common setting~\cite{yan2018second}, we set $ \theta_{o'} = \sin{\theta_o}\cos{\theta_{t}}$ and $\theta_{t'} = \cos{\theta_o}\sin{\theta_{t}}$. ``$\text{Int}$" denotes the intersection volume between the two 4-dimensional bounding boxes. 
$k$ is side length of the 
fourth dimension corresponding to center $\theta$, which we empirically set as 
$k=1$ to achieve the best performance.

Our proposed RDIoU inherits the geometry 
constraints from the 3D IoU through an IoU-like formulation in the 4-dimensional space. Every variable is differentiable
with our RDIoU as the regression target, and thus can be trained with 
the regular gradient-based back-propagation on the GPU. Furthermore, the decoupling of the rotation as an independent variable in our RDIoU leads to the right optimization direction for each individual variable with the increasing IoU during optimization.

\subsection{More Comparison between RDIoU and 3D IoU}
\begin{figure}[t]
	\centering
	\includegraphics[width=1\textwidth]{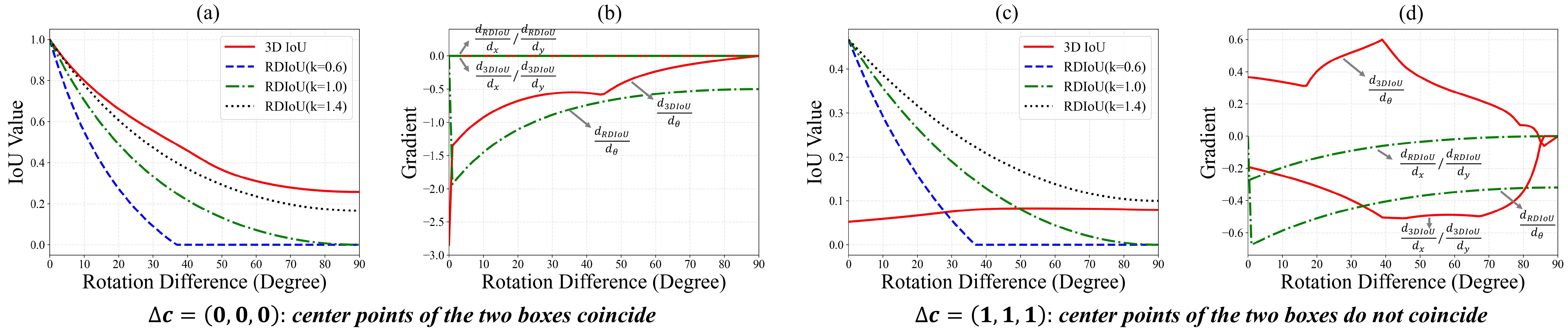} 
	\caption{Simulation experiments. We initialize two 3D boxes of size equal to the predefined anchor size, \textit{i.e.}, (3.9,1.6,1.56). Here, $\Delta{c}$ is the box center distance.
	The gradients of RDIoU are computed with $k=1$.}
	\label{RDIoUvsReal}
\end{figure}

To further investigate the intrinsic difference, we simulate the values of 3D IoU and our RDIoU under increasing rotation differences. As shown in Figure~\ref{RDIoUvsReal}(a), the values of RDIoU and 3D IoU are similar when the center points of the two boxes coincide. 
In contrast, as shown in Figure~\ref{RDIoUvsReal}(c), the values become distinct when the center points of the two boxes do not coincide. Specifically, we can see that the 3D IoU yields unreasonable dynamics in Figure~\ref{RDIoUvsReal}(c), where the computed IoU value increases even when the rotation difference becomes larger. We utilize the hyperparameter $k$ to control the weight of rotation change. As can be seen from Figure~\ref{RDIoUvsReal}(a) and (c), a smaller $k$ results in a steeper curve. Furthermore, we plot the gradient changes of $x,y$ and $\theta$ with increasing rotation difference in Figure~\ref{RDIoUvsReal}(b) and (d) corresponding to Figure~\ref{RDIoUvsReal}(a) and (c), respectively. From Figure~\ref{RDIoUvsReal}(b), we can see that the gradients of 3D IoU and RDIoU are close when the center points of the two boxes coincide. However, as shown in Figure~\ref{RDIoUvsReal}(d), the gradient over rotation of 3D IoU is obviously misdirected ($\frac{d_{3DIoU}}{d_{\theta}} >0$) when the center points of the two boxes do not coincide. In this case, the gradient of 3D IoU changes drastically, while RDIoU still performs in a relatively smooth manner.

\subsection{Incorporating RDIoU into Regression Supervision}
We modify the DIoU loss~\cite{zheng2020distance} for 2D object detection to take our RDIoU for 3D object detection with rotation. 
Specifically, the modified regression loss aims to maximize the intersection volume, \textit{i.e.} the RDIoU and minimize the normalized center distance in the 4-dimensional space.  Let $\bm{c}_{o} = [x_o,y_o,z_o,\theta_{o'}]$ and $\bm{c}_{t} = [x_{t},y_{t},z_{t},\theta_{t'}]$, the center distance penalty is:
\begin{align}
	\rho_c = \frac{\delta(\bm{c}_{o}, \bm{c}_t)}{\text{Diag}},
\end{align}
where $\delta(\cdot)$ denotes the square of Euclidean distance and
\begin{align}
	\nonumber\text{Diag}=& \mathcal{G}(x_o,x_t,l_o,l_t)+\mathcal{G}(y_o,y_t,w_o,w_t)+\mathcal{G}(z_o,z_t,h_o,h_t)+\mathcal{G}(\theta_{o'},\theta_{t'},k,k),
\end{align}
where $\mathcal{G}(a_o,a_t,b_o,b_t) = \big(\max(a_o+\frac{b_o}{2},a_t+\frac{b_t}{2})-\min(a_o-\frac{b_o}{2},a_t-\frac{b_t}{2})\big)^2$. As a result, the final RDIoU-guided DIoU loss for 3D bounding box regression is:
\begin{align}\label{penalty_terms}
	\mathcal{L}_{RL} = 1 - \text{RDIoU} + \rho_c. 
\end{align}

Compared with the existing 3D variants of DIoU losses~\cite{zheng2021se,zhou2019iou} that directly maximize the 3D intersection volume and minimize the 3D center distance, our RDIoU-guided DIoU loss has the advantages of more efficient learning and low resource cost. The rotation decoupling operator transforms the rotated 3D bounding box into a 4-dimensional latent space, which leads to stable optimization direction for each variable. Unlike~\cite{zheng2021se,zhou2019iou} that need huge GPU memory cost or necessary pre-trained model, our RDIoU-guided DIoU loss can be easily implemented in GPU with a small cost. In the ablation studies, we show RDIoU-guided DIoU loss can surpass the 3D IoU-guided DIoU loss by a large margin.

\subsection{Incorporating RDIoU into Classification Supervision}
Solving the misalignment problem is also a crucial issue for object detection. 
Thus, we derive a RDIoU-guided quality focal loss to jointly supervise the classification and the bounding box quality estimation. The quality focal loss~\cite{li2021generalized,li2020generalized} has achieved good performance in 2D image detection, where the IoUs between predicted and ground truth boxes are set as the quality targets. For 3D detection with rotated 3D bounding boxes, an intuitive thought is to apply the quality focal loss with 3D IoU-guided quality supervision. However, 3D IoU can cause confusion when the higher prediction error corresponds to a larger 3D IoU (see Figure~\ref{Real_IoU}), and therefore it leads to bad dynamics during the optimization.

On the contrary, our RDIoU is 
less susceptible to the change of regression variables and thus it can be seen as a 
better measure of the regression quality. Consequently, we utilize our RDIoU as the quality estimation target. Similar to GFLV2~\cite{li2021generalized}, we adopt joint representation targets with the quality focal loss 
for both classification and bounding box quality prediction. Specifically, the target for each sample can be expressed as $\bm{t} = \bm{r}\times e$, where $\bm{r} = [r_1, r_2, \dots, r_m], r_i\in \{0,1\}$ is the classification representation with $m$ classes and $e \in [0,1]$ is the scalar RDIoU representation. 
Taking the joint estimation as $y$, the RDIoU-guided quality focal loss is given by:
\begin{equation}
	\mathcal{L}_{RQFL}  =-\beta_1|\text{RDIoU}-y|^{\beta_2}\big((1-\text{RDIoU})\log(1-y)
	+\text{RDIoU}\log(y)\big),
\end{equation}
where $\beta_1=0.25$ and $\beta_2=2$ follow the setting of focal loss. In this way, the predicted values of classification head 
model both the category probability 
and the corresponding bounding box confidence. Furthermore, we use the cross-entropy loss for direction classification $(\mathcal{L}_{d})$. Finally, the overall loss for training is:
\begin{align}
	\mathcal{L} = \mathcal{L}_{RQFL} + \gamma_1\mathcal{L}_{d} + \gamma_2\mathcal{L}_{RL},
\end{align}
where $\gamma_1$ and $\gamma_2$ are hyperparameters to weigh the 
loss terms.

\noindent\textbf{Backbone Network.}
Following the previous works~\cite{lang2019pointpillars,yan2018second}, we voxelize the point cloud to produce a regular representation. In each voxel, we calculate the mean coordinates and intensities of the raw points as the voxel feature. Subsequently, the 3D backbone network gradually converts the initial voxel features into high-dimensional feature volumes, and the 3D features along Z-axis are stacked into 2D BEV feature maps. 
A 2D backbone network is designed to enhance the extraction of the BEV features for the final location and classification tasks. Motivated by the success of Transformer~\cite{vaswani2017attention} in computer vision-related tasks, 
we adopt a stacked architecture that contains both the convolution block and the Transformer block to generate the final detection features. Specifically, the convolution block first extracts the BEV features while keeping the dimensions (both the number of channels and the size of feature map) unchanged. 
A convolution layer with stride two is then adopted to further reduce the spatial size. The Transformer block then processes the high-level features without changing the feature dimensions. Our designed conv-trans-stacked (CT-stacked) architecture shows better performance as compared to the pure convolution based network. The whole framework of our RDIoU is illustrated in Figure~\ref{model}. The final feature map is $8\times$ downsampling of the input resolution. The classification, quality estimation, and regression are performed at pixel-level. 

\section{Experiments}
In this section, we compare our RDIoU to other state-of-the-art methods on two popular benchmarks: KITTI~\cite{geiger2013vision} and Waymo Open Dataset~\cite{sun2020scalability}. Furthermore, we conduct extensive ablation studies to investigate the performance on different backbone networks and on each component of RDIoU to validate our design.

\begin{table}[tp]  
	\tiny
	\setlength\tabcolsep{2pt}
	\begin{center}
		\resizebox{\textwidth}{!}
		{
			\begin{tabular}{c|c|c||ccc|ccc|ccc}
				\toprule
				\multirow{2}*{Type}&\multirow{2}*{Method}&\multirow{2}*{M}& \multicolumn{3}{c|}{Car-3D (IoU=0.7)} & \multicolumn{3}{c|}{Ped.-3D (IoU=0.5)} & \multicolumn{3}{c}{Cyc.-3D (IoU=0.5)}\\
				&&& Easy & Mod.* & Hard & Easy & Mod. &Hard  & Easy & Mod. &Hard\\
				\midrule
				\multirow{8}*{\rotatebox{90}{Two-stage}}&Part-$A^2$~\cite{shi2020points}&\Checkmark &\textbf{89.56}&79.41&78.84&\textbf{65.69}& \textbf{60.05}&55.45&85.50&69.90&65.49\\
				&STD~\cite{yang2019std}&\XSolidBrush &89.70&79.80&\textbf{79.30}&-&-&-&-&-&-\\
				&PV-RCNN~\cite{shi2020pv}&\Checkmark &89.35& 83.69& 78.70&63.12&54.84&51.78& \textbf{86.06}&69.48&64.50\\
				&Voxel-RCNN~\cite{deng2020voxel}&\XSolidBrush &89.41& 84.52& 78.93&-&-&-&-&-&-\\
				&VoTr-TSD~\cite{mao2021voxel} &\XSolidBrush &89.04 &84.04 &78.68&-&-&-&-&-&-\\
				&CT3D~\cite{sheng2021improving} &\Checkmark &89.11& 85.04& 78.76&64.23& 59.84& \textbf{55.76}&85.04&\textbf{71.71}&\textbf{68.05} \\
				&CT3D~\cite{sheng2021improving} &\XSolidBrush & 89.54&86.06&78.99&-&-&-&-&-&- \\
				&BtcDet~\cite{xu2021behind}&\XSolidBrush &- & \textbf{86.57}&-&-&-&-&-&- \\
				\midrule
				\midrule
				\multirow{12}*{\rotatebox{90}{Single-stage}}&VoxelNet~\cite{zhou2018voxelnet}&\Checkmark &81.97 &65.46 &62.85 &57.86& 53.42& 48.87& 67.17 &47.65 &45.11\\
				&SECOND~\cite{yan2018second}&\Checkmark &88.61& 78.62&77.22 &56.55&52.98&47.73&80.59&67.16&63.11\\
				&PointPillar~\cite{lang2019pointpillars}&\Checkmark &86.46&77.28&74.65&57.75&52.29&47.91&80.06&62.69& 59.71\\
				&3DSSD~\cite{yang20203Dssd}&\Checkmark &89.71&79.45& 78.67&-&-&-&-&-&-\\
				&SA-SSD~\cite{he2020structure}&\XSolidBrush&\textbf{90.15}&79.91&78.78&-&-&-&-&-&-\\
				&CIA-SSD~\cite{zheng2020cia}&\XSolidBrush& 90.04&78.91&78.80&-&-&-&-&-&-\\
				&SE-SSD~\cite{zheng2021se}&\XSolidBrush&-&85.71&-&-&-&-&-&-&-\\
				&VoTr-SSD~\cite{mao2021voxel} &\XSolidBrush&87.86 &78.27& 76.93&-&-&-&-&-&-\\
				\cline{2-12}
				&\textbf{RDIoU (Ours)}&\Checkmark &89.16&85.24&78.41&\textbf{63.26}&\textbf{57.47}&\textbf{52.53}&\textbf{83.32}& \textbf{68.39}& \textbf{63.63}\\
				&\textbf{RDIoU (Ours)}&\XSolidBrush &89.76&\textbf{86.62}&\textbf{79.04}&-&-&-&-&-&-\\
				\bottomrule
			\end{tabular}
		}
	\end{center}
	\caption{Performance comparisons with state-of-the-art methods on the KITTI \textit{val} set with 11 recall positions. $\text{M}:$\Checkmark means training on three classes. $\text{M}:$\XSolidBrush means training only on car. Mod.* is the most important metric. The top-1 of two-stage and single-stage methods are bold, respectively.} 
	\label{table:kitti_val}
\end{table}

\begin{table}[t]  
	\tiny
	\setlength\tabcolsep{4pt}
	\begin{center}
		\resizebox{\textwidth}{!}
		{
			\begin{tabular}{c|c|c||ccc}
				\toprule
				\multirow{2}*{Method} &\multirow{2}*{Reference}&\multirow{2}*{Stage}& \multicolumn{3}{c}{3D AP (IoU=0.7)}  \\
				&&& Easy & Moderate* & Hard  \\
				\midrule
				Part-$A^2$~\cite{shi2020points}&TPAMI 2020&Two&87.81& 78.49& 73.51\\
				STD~\cite{yang2019std}&ICCV 2019&Two&87.95& 79.71& 75.09\\
				Point-GNN~\cite{shi2020pointgnn}&CVPR2020&Two&88.33& 79.47& 72.29\\
				PV-RCNN~\cite{shi2020pv}&CVPR 2020&Two&90.25& 81.43& 76.82\\
				LiDAR-RCNN~\cite{li2021lidar}&CVPR 2021&Two&85.97&74.21&69.18\\
				VoTr-TSD~\cite{mao2021voxel}&ICCV 2021&Two&89.90 &82.09 &\textbf{79.14} \\
				CT3D~\cite{sheng2021improving}&ICCV 2021&Two&87.83& 81.77 &77.16\\
	        	BtcDet~\cite{xu2021behind}&AAAI 2022& Two & \textbf{90.64} & \textbf{82.86} &78.09\\
				\midrule
				VoxelNet~\cite{zhou2018voxelnet}&CVPR 2018&Single&77.82& 64.17& 57.51\\
				SECOND~\cite{yan2018second}& Sensors 2018&Single&83.34& 72.55& 65.82\\
				PointPillar~\cite{lang2019pointpillars}&CVPR 2019&Single&82.58& 74.31& 68.99\\
				3DSSD~\cite{yang20203Dssd}&CVPR 2020&Single&88.36& 79.57& 74.55\\
				SA-SSD~\cite{he2020structure}&CVPR 2020&Single&88.75& 79.79& 74.16\\
				SE-SSD~\cite{zheng2021se}&CVPR 2021&Single&\textbf{91.49}& \textbf{82.54}& 77.15\\
				\midrule
				\textbf{RDIoU (Ours)}&-&Single&90.65&82.30&\textbf{77.26}\\
				\bottomrule
			\end{tabular}
		}
	\end{center}
	\caption{Performance comparisons with state-of-the-art methods for car detection on the KITTI \textit{test} benchmark with 40 recall positions. The top-1 of two-stage and single-stage methods are bold, respectively.} 
	\label{table:kitti_test}
\end{table}

\subsection{Datasets}
\noindent\textbf{KITTI.} This dataset consists of 7,481 LiDAR samples for training and 7,518 LiDAR samples for testing. We further follow the common protocol~\cite{lang2019pointpillars,yan2018second} to split the original training data into 3,712 training samples and 3,769 validation samples for experimental studies. 

\noindent\textbf{Waymo Open Dataset.} This dataset consists of 798 training sequences with 158,361 LiDAR samples, and 202 validation sequences with 40,077 LiDAR samples. Five LiDAR sensors are used for full 360-degree annotation instead of 90-degree as in KITTI. Currently, it is the largest dataset for autonomous driving. 

\subsection{Implementation Details}
For KITTI, the raw point clouds are first clipped into $(0, 70.4)$m, $(-40, 40)m$, $(-3,1)$m for $X, Y, Z$ axis ranges with voxel size $(0.05, 0.05, 0.1)$m. For Waymo Open Dataset, the corresponding axis ranges are $(-75.2, 75.2)$m, $(-75.2, 75.2)$m, $(-2, 4)$m, and the voxel size is $(0.1, 0.1, 0.15)$m. We conduct all experiments based on the OpenPCDet~\cite{openpcdet2020} toolbox.

The 3D backbone has four levels with feature dimensions $(16, 32, 64, 64)$, $(16, 32, 64, 128)$ for KITTI and Waymo Open Dataset, respectively. The 2D backbone contains two blocks, the first block is implemented by 5 CNN layers for KITTI and 6 CNN layers for Waymo Open Dataset to keep the same resolution with the output of 3D backbone. The second block is implemented by 1 CNN layer and 4 Transformer (using the setting of 4$\times$ expansion in FFN layers and 4 attention heads) layers~\cite{liu2021swin,vaswani2017attention} with half the resolution. Finally, one fractionally-strided convolution layer is adopted to double the resolution. 

All the evaluated models are trained from scratch in an end-to-end manner with the ADAM optimizer. The learning rate is decayed with a cosine annealing strategy, and the maximum is $3.5\times e^{-4}$. We use a batch size of 32. For the hyperparameters of the overall loss, we set $\gamma_1=0.2$ and $\gamma_2=2$.

\begin{table*}[tp]  
	\setlength\tabcolsep{4pt}
	\begin{center}
		\resizebox{\textwidth}{!}
		{
			\begin{tabular}{c|c||cccc|cccc}
				\toprule
				\multirow{2}*{Method} &\multirow{2}*{Stage} & \multicolumn{4}{c|}{3D AP/APH (IoU=0.7)} &  \multicolumn{4}{c}{BEV AP/APH (IoU=0.7)}\\ 
				&& Overall & 0-30m & 30-50m & 50m-Inf & Overall & 0-30m & 30-50m & 50m-Inf\\
				\midrule
				\multicolumn{10}{c}{\textit{\textbf{LEVEL\_1}}}\\
				\midrule
				MVF~\cite{zhou2020end} &Two & 62.9/- & 86.3/- & 60.0/- & 36.0/- &80.4/-&93.6/-&79.2/-&63.1/-\\
				PV-RCNN~\cite{shi2020pv} &Two& 70.3/69.7 & 91.9/91.3 & 69.2/68.5 & 42.2/41.3 &80.0/82.1& 97.4/96.7& 83.0/82.0 &65.0/63.2\\
				Voxel-RCNN~\cite{deng2020voxel} &Two& 75.6/- & 92.5/- & 74.1/- & 53.2/-&88.2/-& 97.6/-& 87.3/-& 77.7/-\\
				LiDAR-RCNN~\cite{li2021lidar} &Two&76.0/75.5& 92.1/91.6&74.6/74.1 &54.5/53.4&90.1/89.3&97.0/96.5&\textbf{89.5/88.6}& \textbf{78.9/77.4}\\
				CenterPoint~\cite{yin2021center} &Two&76.7/76.2&-/-&-/-&-/-&&-/-&-/-&-/-\\
				VoTr-TSD~\cite{mao2021voxel}&Two&75.0/74.3&92.3/91.7&73.4/72.6&51.1/50.0&-/-&-/-&-/-&-/-\\
				CT3D~\cite{sheng2021improving}&Two&76.3/-&92.5/-& 75.1/-&\textbf{55.4/-}&\textbf{90.5/-}&\textbf{97.6/-}&88.1/-&78.9/-\\
				BtcDet~\cite{xu2021behind}&Two&\textbf{78.6/78.1}&\textbf{96.1/-}&\textbf{77.6/-}&54.5/-&-/-&-/-&-/-&-/-\\
				\hline
				PointPillar*~\cite{lang2019pointpillars}&Single  & 72.1/71.5& 88.3/87.8& 69.9/69.3& 48.0/47.3 &87.9/87.1& 96.6/96.0& 87.1/86.2& 78.1/76.5 \\
				Pillar-OD~\cite{wang2020pillar} &Single& 69.8/- & 88.5/- & 66.5/- & 42.9/- &87.1/-&95.8/-&84.7/-&72.1/-\\
				VoTr-SSD~\cite{mao2021voxel}&Single& 69.0/68.4&88.2/87.6&66.7/66.1&42.1/41.4&-/-&-/-&-/-&-/-\\
				\textbf{RDIoU (Ours)} &Single &\textbf{78.4/78.0}&\textbf{93.0/92.6}&\textbf{75.4/74.9}&\textbf{56.2/55.6}&\textbf{91.6/91.0}&\textbf{98.1/97.7}&\textbf{90.8/90.2}&\textbf{82.4/81.1}\\
				\midrule
				\multicolumn{10}{c}{\textit{\textbf{LEVEL\_2}}}\\
				\midrule
				PV-RCNN~\cite{shi2020pv}&Two  & 65.4/64.8 & 91.6/91.0 & 65.1/64.5 & 36.5/35.7 &77.5/76.6& 94.6/94.0& 80.4/79.4& 55.4/53.8\\
				Voxel-RCNN~\cite{deng2020voxel}&Two & 66.6/- & 91.7/- & 67.9/- & 40.8/-&81.1/-& 97.0/-& 81.4/-& 63.3/-\\
				LiDAR-RCNN~\cite{li2021lidar}&Two &68.3/67.9&91.3/90.9& 68.5/68.0& 42.4/41.8&81.7/81.0&94.3/93.9&\textbf{82.3}/\textbf{81.5}&\textbf{65.8}/\textbf{64.5}\\
				CenterPoint~\cite{yin2021center}&Two &68.8/68.3&-/-&-/-&-/-&-/-&-/-&-/-&-/-\\
				VoTr-TSD~\cite{mao2021voxel}&Two&65.9/65.3&-/-&-/-&-/-&-/-&-/-&-/-&-/-\\
				CT3D~\cite{sheng2021improving}&Two&69.0/-&91.8/-&68.9/-&42.6/-&\textbf{81.7/-}&\textbf{97.1/-}&82.2/-&64.3/-\\
				BtcDet~\cite{xu2021behind}&Two&\textbf{70.1/69.6}&\textbf{96.0/-}&\textbf{70.1/-}&\textbf{43.9/-}&-/-&-/-&-/-&-/-\\
				\hline
			    PointPillars*~\cite{lang2019pointpillars}&Single&63.6/63.1 & 87.4/86.9 &62.9/62.3& 37.2/36.7& 81.3/80.4& 94.0/93.5 &81.7/80.8& 65.5/64.1\\
				VoTr-SSD~\cite{mao2021voxel}&Single&60.2/59.7&-/-&-/-&-/-&-/-&-/-&-/-&-/-\\
				\textbf{RDIoU (Ours)}&Single &\textbf{69.5/69.1}&\textbf{92.3/91.9}&\textbf{69.3/68.9}&\textbf{43.7/43.1}&\textbf{83.1/82.5}&\textbf{97.5/97.1}&\textbf{85.2/84.6}&\textbf{68.3/67.2}\\
				\bottomrule
			\end{tabular}
		}
	\end{center}
	\caption{Performance comparisons with state-of-the-art methods for the vehicle detection on the Waymo validation dataset. Here PointPillars* is implemented in mmdetection3D~\cite{mmdet3d2020}. The top-1 of two-stage and single-stage methods are bold, respectively.} 
	\label{table:waymo_val}
\end{table*}

\subsection{Results on Real-world Datasets}
\textbf{KITTI.} KITTI is relatively small as compared to the Waymo Open Dataset. We follow the previous works to train models on \textit{train} set, and report the results on \textit{val} set. Furthermore, we report the detection results on KITTI \textit{test} server by training the model with \textit{train+val} set. All the evaluated models are reported 
in three difficulty levels (\textit{i.e.}, \textit{easy}, \textit{moderate}, \textit{hard}). 
Table \ref{table:kitti_val} reports the results on the KITTI \textit{val} set. For the most important 3D object detection metric on \textit{moderate} level of car, our proposed RDIoU method surpasses the current best single-stage models CIA-SSD~\cite{zheng2020cia}, SE-SSD~\cite{zheng2021se}, VoTr-SSD~\cite{mao2021voxel} with +7.71\%AP, +0.91\%AP, +8.35\%AP, respectively. Furthermore, our RDIoU even outperforms the best two-stage models VoTr-TSD~\cite{mao2021voxel}, CT3D~\cite{sheng2021improving} BtcDet~\cite{xu2021behind} with +2.58\%AP, +0.56\%AP, +0.05\%AP, respectively. This superior performance strongly manifests the effectiveness of our proposed method. 

For pedestrian and cyclist detection, we perform a three-class training for our RDIoU method and compare to the state-of-the-art methods with the same settings. It can be seen that our RDIoU leads a large margin compared with these single-stage methods (VoxelNet~\cite{zhou2018voxelnet}, SECOND~\cite{yan2018second} and PointPillar~\cite{lang2019pointpillars}). At the \textit{moderate} level, our RDIoU outperforms SECOND with +4.49\%AP and +1.23\%AP on pedestrian and cyclist detection, respectively. This also proves that our RDIoU has the superior ability of detecting small objects. Note that the performances on small objects are marginally lower compared to the other two-stage methods such as Part-A$^2$~\cite{shi2020points} and CT3D~\cite{sheng2021improving}. This is because single-stage methods usually assign more attention to the main objects (\textit{i.e.,} car) compared to the two-stage methods which have the inherent advantage from the proposals.

Table~\ref{table:kitti_test} shows the car-3D detection results by submitting to the KITTI \textit{test} server. Our RDIoU achieves 82.30\%AP on the most important \textit{moderate} level, surpassing the state-of-the-art methods of LiDAR-RCNN~\cite{li2021lidar}, VoTr-TSD~\cite{mao2021voxel}, CT3D~\cite{sheng2021improving} by +8.09\%AP, +0.21\%AP, +0.53\%AP, respectively. Note that these are all two-stage methods such that RDIoU has the absolute advantage of inference speed. Compared to SE-SSD~\cite{zheng2021se} and BtcDet~\cite{xu2021behind}, our RDIoU is slightly lower by -0.24\%AP and -0.56\%AP, respectively. One 
possible reason is the mismatched data distributions between the KITTI \textit{val} set and \textit{test} set~\cite{li2021lidar,shi2020points}. Moreover, SE-SSD requires extra data augmentation and complex self-ensembling procedure while the two-stage method BtcDet requires a further box refinement module. Overall, the results on both \textit{test} and \textit{val} sets consistently reveal that RDIoU method is highly effective.

\noindent\textbf{Waymo Open Dataset.} All 
methods for comparison are evaluated with AP and average precision by heading (APH) at two difficulty levels defined in the official evaluation, where the \textit{LEVEL\_1} objects contain at least 5 points while the \textit{LEVEL\_2} objects contain 1$\sim$5 inside points. The rotated IoU threshold is set to 0.7 for vehicle detection. We also report the detection results based on the different distances of the objects for adequate comparison.
Table~\ref{table:waymo_val} reports the results on validation sequences of Waymo Open Dataset. It can be clearly seen that RDIoU achieves excellent performance on this large-scale, diverse, and challenging dataset. Our RDIoU beats almost all state-of-the-arts including the two-stage methods on all evaluation metrics by a large margin in a number of settings. Specifically, our RDIoU outperforms the latest state-of-the-art methods CenterPoint~\cite{yin2021center}, VoTr-TSD~\cite{mao2021voxel}, CT3D~\cite{sheng2021improving} with +1.7\%, +3.4\%, +2.1\% on 3D AP of \textit{LEVEL\_1}, respectively, and achieves fairly close performance as compared to BtcDet~\cite{xu2021behind} (only -0.2\%AP performance drop). These inspired results further affirm the strong ability of our RDIoU method.

\begin{table}[t]
	\scriptsize
	\setlength\tabcolsep{4.5pt}
	\begin{center}
		\begin{tabular}{c||ccc|c|c}
			\toprule
			\multirow{2}*{Method}& \multicolumn{3}{c|}{$\text{3D}_{R11}$} & $\text{3D}_{R40}$&FPS\\
			& Easy & Mod. & Hard & Mod.&(Hz)\\
			\midrule
			PointPillar~\cite{lang2019pointpillars} &87.08&77.74&76.24&79.88&33.8\\
			PointPillar (+RDIoU) &\textbf{88.89}&\textbf{78.89}&\textbf{78.02}&\textbf{82.42}&33.8\\
			\midrule
			SECOND~\cite{yan2018second}&88.78&78.74&77.51&82.85&30.5\\
			SECOND (+RDIoU)&\textbf{89.24}&\textbf{86.10}&\textbf{78.60}&\textbf{85.80}&30.5\\
			\midrule
			CT-stacked& 88.93&78.91&77.63&83.01&26.6\\
			CT-stacked (+RDIoU)& \textbf{89.76}&\textbf{86.62}&\textbf{79.04}&\textbf{86.20}&26.6\\
			\bottomrule
		\end{tabular}
	\end{center}
	\caption{Ablation study on different backbone networks. (+RDIoU) means replacing the classification and regression losses in baseline models with our proposed RDIoU.} 
	\label{table:abla_backbone}
\end{table}

\subsection{Ablation Studies}
In this section, we detail the influence of each component that contributes to the final RDIoU design. All models are trained on the KITTI \textit{train} set from scratch, and evaluated on the KITTI \textit{val} set for fair comparison. We use CT-stacked backbone network as the default setting. The final performance is reported on 3D \textit{moderate} level of car with 11 and 40 recall positions. 

\noindent\textbf{Effect of Different Backbone Networks.} In Table~\ref{table:abla_backbone}, we plug RDIoU into another two commonly used 3D object detection backbone networks SECOND~\cite{yan2018second} and PointPillar~\cite{lang2019pointpillars}. It can be seen that RDIoU can help different backbone networks to achieve significant performance improvement. Specifically, plugging our RDIoU into PointPillar, SECOND and our CT-stacked backbone networks brings $\text{3D}_{R40}$+2.54\%, $\text{3D}_{R40}$+2.95\% and $\text{3D}_{R40}$+3.19\% improvements on \textit{moderate} level, respectively. 
The contributing factor is that our RDIoU method can be integrated into any existing backbone networks to assist the model learning for better performance.
\begin{table}[t]
	\scriptsize
	\setlength\tabcolsep{4.5pt}
	\begin{center}
		\begin{tabular}{c|cc||ccc|c}
			\toprule
			\multirow{2}*{Method} & 3D IoU & RDIoU &\multicolumn{3}{c|}{$\text{3D}_{R11}$} & $\text{3D}_{R40}$ \\
			& guided & guided & Easy & Mod. & Hard & Mod.\\
			\midrule
			Smooth$-\ell1$ loss~\cite{lin2017focal}&& &89.19 &78.86 &77.53 &83.14\\
			\midrule
			\multirow{2}*{IoU loss~\cite{zhou2019iou}} &\checkmark&&88.80&81.94&77.67&83.60\\
			&&\checkmark&\textbf{89.40}& \textbf{85.60}&\textbf{78.76}&\textbf{85.73}\\
			\midrule
			\multirow{2}*{CIoU loss~\cite{zheng2020distance}} &\checkmark&& 88.82&83.20&77.48&84.01\\
			&&\checkmark&\textbf{89.43} & \textbf{86.21} & \textbf{78.90}&\textbf{85.98}\\
			\midrule
			\multirow{2}*{DIoU loss~\cite{zheng2020distance}} &\checkmark&&88.83 &83.59&77.93&84.20 \\
			&&\checkmark&\textbf{89.76}&\textbf{86.62}&\textbf{79.04}&\textbf{86.20}\\
			\bottomrule
		\end{tabular}
	\end{center}
	\caption{Ablation study on RDIoU-guided regression loss. All models utilize RDIoU-guided quality focal loss for classification.} 
	\label{table:abla_RDIoU}
\end{table}

\begin{table}[t]
	\scriptsize
	\setlength\tabcolsep{4.5pt}
	\begin{center}
		\begin{tabular}{cc||ccc|c}
			\toprule
			 3D IoU&RDIoU&\multicolumn{3}{c|}{$\text{3D}_{R11}$} & $\text{3D}_{R40}$ \\
			guided QFL&guided QFL& Easy & Mod. & Hard & Mod.\\
			\midrule
			&&89.16&84.78& 78.15&85.12\\
			\checkmark&&89.33 & 86.07 & 78.79 & 85.79\\
			&\checkmark&\textbf{89.76}&\textbf{86.62}&\textbf{79.04}&\textbf{86.20}\\
			\bottomrule
		\end{tabular}
	\end{center}
	\caption{Ablation study on RDIoU-guided classification loss. Traditional focal loss is adopted as the default setting. All models utilize RDIoU-guided DIoU loss.} 
	\label{table:abla_qfl}
\end{table}

\noindent\textbf{Effect of RDIoU-guided Regression Branch.}
As shown in Figure~\ref{table:abla_RDIoU}, although 3D IoU-based regression losses show better performance than the Smooth-$\ell1$ loss, our RDIoU-based regression losses can still further boost the performance by a large margin. Specifically, RDIoU-guided IoU loss~\cite{zhou2019iou}, CIoU loss~\cite{zheng2020distance}, DIoU loss~\cite{zheng2020distance} surpass the 3D IoU-guided losses with $\text{3D}_{R40}$+2.13\%, $\text{3D}_{R40}$+1.97\%, $\text{3D}_{R40}$+2.00\%, respectively. These gains in AP strongly prove the effectiveness of our proposed method, which encourages the network to learn more precise 3D bounding boxes. DIoU loss performs better than IoU loss due to the direct minimization for the normalized distance between predicted and ground truth bounding boxes. Note that the aspect ratio of the 3D real-world category is relatively stable, and thus CIoU loss does not bring performance improvement as compared to the DIoU loss. 

\noindent\textbf{Effect of RDIoU-guided Classification Branch.}
In Table~\ref{table:abla_qfl}, we investigate the influence of RDIoU-guided quality focal loss by replacing it with the traditional focal loss~\cite{lin2017focal} and the 3D IoU-guided quality foal loss~\cite{li2020generalized}, respectively. As seen in $1^{st}$ and $2^{nd}$ rows of Table~\ref{table:abla_qfl}, the 3D IoU-guided quality focal loss performs better than the traditional focal loss due to the extra quality estimation for each predicted bounding box. $2^{nd}$ row and $3^{rd}$ row of Table~\ref{table:abla_qfl} demonstrate that RDIoU-guided quality focal loss can also obviously exceed 3D IoU-guided quality focal loss ($\text{3D}_{R11}$+0.55\%, $\text{3D}_{R40}$+0.41\%). This significant improvement comes from the stable measurement of the predicted bounding boxes, which subsequently facilitates the generation of more reliable confidence maps.

\section{Conclusion}
This paper presents a novel RDIoU method to model the intersection of two arbitrary-oriented 3D bounding boxes for improving the single-stage point cloud detectors. RDIoU decouples the rotation variable as an independent term while preserving the geometry of 3D IoU. It exhibits robustness to the rotation change compared with the 3D IoU. Based on RDIoU, we propose the RDIoU-guided DIoU loss to enable stable optimization process and efficient back-propagation during training, and it achieves the best performance among the latest 3D regression losses. Moreover, we introduce the RDIoU-guided quality focal loss to address the misalignment problem between the regression results and the confidence maps, and it also exhibits much better than the 3D IoU-guided quality focal loss. The experimental results show that our RDIoU method can help the commonly used backbone networks to achieve state-of-the-art performance. 
\\
\\
\textbf{Acknowledgements.} This work was supported by the National Key R$\&$D Program of China under Grant 2020AAA0103902; Zhejiang Provincial Key Laboratory of Information Processing, Communication and Networking (IPCAN), Hangzhou 310027, China; Fundamental Research Funds for the Central Universities 226-2022-00195; 
The National Research Foundation, Singapore under its AI Singapore Programme (AISG Award No: AISG2-RP-2021-024); The Tier 2 grant MOE-T2EP20120-0011 from the Singapore Ministry of Education.

\clearpage
\bibliographystyle{splncs04}
\bibliography{eccv2022submission.bbl}

\newpage
\appendix

\section{Appendix}
\noindent Some supplementary materials are provided to further validate the proposed approach. The supplementary materials include ablation studies on parameter $k$, more comparisons to existing ODIoU loss~\cite{zheng2021se} and 3D IoU loss~\cite{zhou2019iou}, inserting RDIoU into state-of-the-art 3D detectors, and visualization results, respectively.
\\

\noindent\textbf{Effect of parameter $k$.}  As shown in Table~\ref{table:abla_k}, we try different settings of the edge $k$ to control the weight of rotation change. Here, we can see that $k$ is an important factor to boost the final performance, and $k=1.0$ is the best. 
\begin{table}[h]
	\scriptsize
	\setlength\tabcolsep{4.5pt}
	\begin{center}
		\begin{tabular}{c||c|c|c|c|c|c|c}
			\toprule
			$k$ &0.6 & 0.8 & \textbf{1.0} & 1.2 & 1.4 & 1.6 & 1.8\\
			\midrule
			$\text{3D}_{R40}$&85.89&85.96&\textbf{86.20}&85.92&85.85&85.70&85.59\\
			\bottomrule
		\end{tabular}
	\end{center}
	\caption{Ablation study on parameter $k$, which reflects the sensitivity of RDIoU value to the rotation difference.} 
	\label{table:abla_k}
\end{table}

\noindent \textbf{More comparison with 3D IoU~\cite{zhou2019iou} and ODIoU~\cite{zheng2021se}.} In Table~\ref{table:compare_iou}, we provide more comparison results for 3D IoU and ODIoU with our proposed RDIoU. It can be seen that RDIoU outperforms both 3D IoU and ODIoU by large margins.
\begin{table}[h]
    \setlength\tabcolsep{10pt}
	\begin{center}
		\begin{tabular}{c|ccc|c}
			\toprule
			\multirow{2}*{Method}& \multicolumn{3}{c|}{$\text{3D}_{R11}$} & $\text{3D}_{R40}$\\
			& Easy & Mod. & Hard & Mod.\\
			\midrule
			PointPillar\cite{lang2019pointpillars} &87.08&77.74&76.24&79.88\\
			PointPillar  (+3D IoU) & 86.91 & 77.93 & 76.83 & 80.12\\
			PointPillar  (+ODIoU) & 87.15 & 78.29 & 77.08 & 80.58\\
			PointPillar (+RDIoU) &\textbf{88.89}&\textbf{78.89}&\textbf{78.02}&\textbf{82.42}\\
			\midrule
			SECOND\cite{yan2018second}&88.78&78.74&77.51&82.85\\
		    SECOND  (+3D IoU) & 88.04 & 80.97 & 77.06 & 83.02\\
			SECOND  (+ODIoU) & 88.69 & 82.82 & 77.41 & 83.88\\
			SECOND (+RDIoU)&\textbf{89.24}&\textbf{86.10}&\textbf{78.60}&\textbf{85.80}\\
			\midrule
			CT-stacked& 88.93&78.91&77.63&83.01\\
			CT-stacked (+3D IoU) & 88.23 & 81.09 & 77.16 & 83.29\\
			CT-stacked (+ODIoU) & 88.70 & 82.89 & 77.53 & 84.01\\
			CT-stacked (+RDIoU)& \textbf{89.76}&\textbf{86.62}&\textbf{79.04}&\textbf{86.20}\\
			\bottomrule
		\end{tabular}
	\end{center}
 	\caption{Comparisons to 3D IoU and ODIoU.} 
    \label{table:compare_iou}
\end{table}

\noindent \textbf{Inserting RDIoU into state-of-the-art methods.} We reproduce SOTA single-stage methods: CIA-SSD\cite{zheng2020cia}, SA-SSD\cite{he2020structure}, SE-SSD\cite{zheng2021se}, and two-stage methods: PV-RCNN\cite{shi2020pv}, Voxel-RCNN\cite{deng2020voxel}, CT3D\cite{sheng2021improving} in 
Table~\ref{table:sota}. Our RDIoU can significantly improve their performance, especially for the single-stage detectors.
\begin{table}[h]  
    \vspace{-3mm}
    \setlength\tabcolsep{10pt}
	\begin{center}
		\begin{tabular}{c|c|ccc|c}
			\toprule
			\multirow{2}*{Type}&\multirow{2}*{Method}& \multicolumn{3}{c|}{$\text{3D}_{R11}$}&{$\text{3D}_{R40}$}\\
			&& Easy & Mod. & Hard & Mod.\\
		    \midrule
			\multirow{6}*{\rotatebox{90}{Single-stage}} &CIA-SSD\cite{zheng2020cia} & \textbf{89.48} & 78.54 & 77.35 & 81.93\\
			&CIA-SSD (+RDIoU)&89.07&\textbf{85.44}&\textbf{78.55}&\textbf{85.23}\\
            \cline{2-6}
			&SA-SSD\cite{he2020structure}&89.26 & 79.28 & 78.35 & 82.65\\
			&SA-SSD (+RDIoU)&\textbf{89.56}&\textbf{86.04}&\textbf{78.76}&\textbf{85.87}\\
            \cline{2-6}
			&SE-SSD\cite{zheng2021se} & 89.07 & 79.22 & 78.37 & 82.48\\
			&SE-SSD (+RDIoU)& \textbf{89.24}&\textbf{85.98}&\textbf{78.60}&\textbf{85.24}\\
			\midrule
			\multirow{6}*{\rotatebox{90}{Two-stage}} & PV-RCNN\cite{shi2020pv} & 89.31 & 84.49 & 78.78& 84.93\\
			& PV-RCNN (+RDIoU) & \textbf{89.47} & \textbf{86.21} & \textbf{79.01} & \textbf{85.96}\\
			\cline{2-6}
			& Voxel-RCNN\cite{deng2020voxel} & 89.44 & 84.45 & 78.90 & 85.24\\
			& Voxel-RCNN (+RDIoU) &  \textbf{89.67} & \textbf{86.12} & \textbf{78.91} & \textbf{85.92}\\
			\cline{2-6}
			& CT3D\cite{sheng2021improving} & \textbf{89.54} & 86.06 & \textbf{78.99} & 85.82\\
			& CT3D (+RDIoU) & 89.31 & \textbf{86.27} & 78.90 & \textbf{85.94}\\
			\bottomrule
		\end{tabular}
	\end{center}
	\caption{Inserting our RDIoU into SOTA methods based on OpenPCDet\cite{openpcdet2020}.} 
	\label{table:sota}
    \vspace{-3mm}
\end{table}

\begin{figure}[h]
    \vspace{-3mm}
	\centering
	\includegraphics[width=1.0\textwidth]{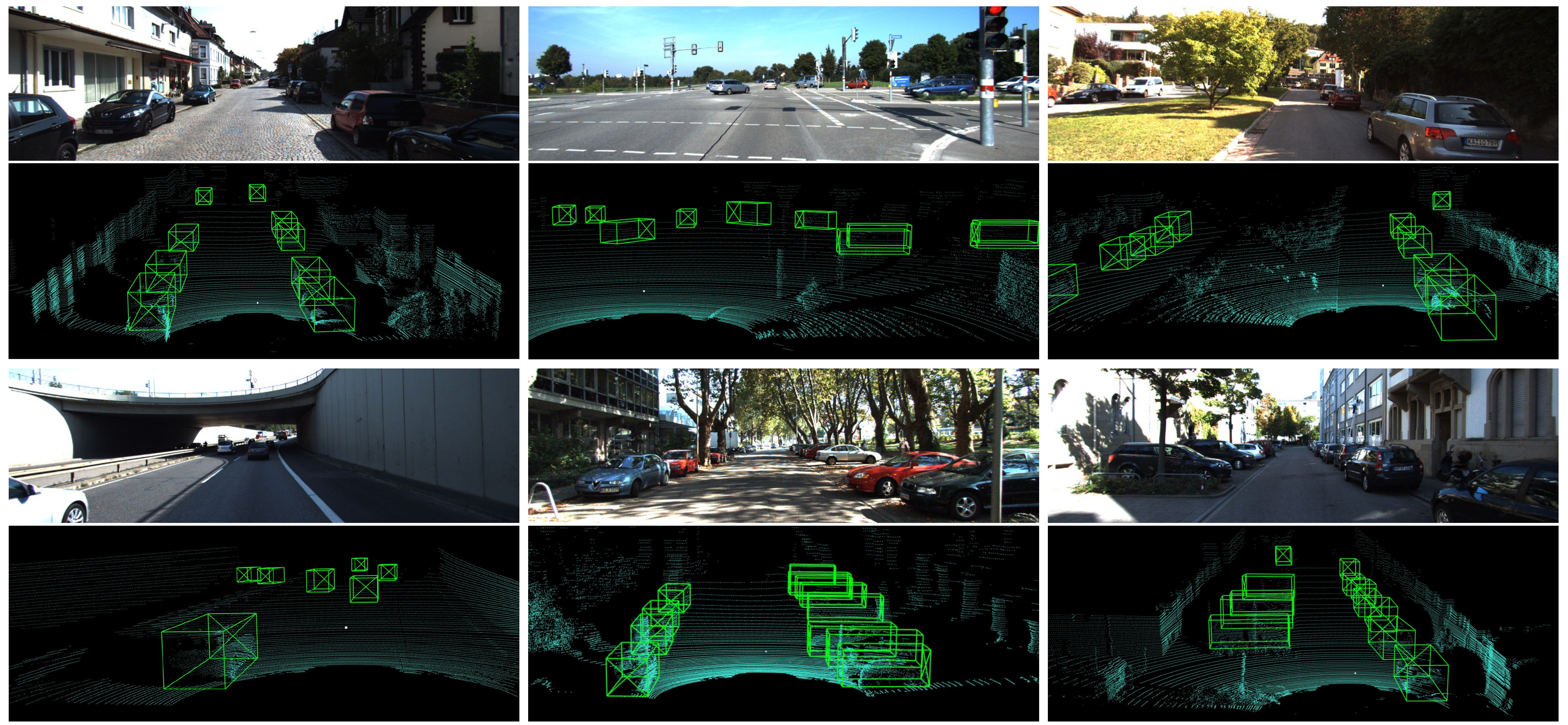}
    \caption{Snapshots of qualitative results on KITTI test set. The output 3D bounding boxes are shown in green.}
    \label{fig:vis}
\end{figure}

\noindent \textbf{Visualization.} We provide some qualitative detection results on KITTI \textit{test} set in Figure~\ref{fig:vis}, it shows that RDIoU can produce high-quality 3D bounding boxes in diversified scenarios.

\end{document}